\documentclass[11pt,a4paper]{article}
\usepackage[hyperref]{emnlp2018}
\usepackage{times}
\usepackage{latexsym}
\usepackage{graphicx}
\usepackage{multirow}
\usepackage{todonotes}
\usepackage[utf8]{inputenc}

\usepackage{url}

\aclfinalcopy 

\title{Revisiting Character-Based Neural Machine Translation\\with Capacity and Compression}
\author{
    Colin Cherry\thanks{*Equal contributions}, $\phantom{ }$
    George Foster\footnotemark[1], $\phantom{ }$
    Ankur Bapna, $\phantom{ }$
    Orhan Firat, $\phantom{ }$
    Wolfgang Macherey \\
  Google AI \\
  {\tt colincherry,fosterg,ankurbpn,orhanf,wmach@google.com}}
\date{}

\begin{document}
\maketitle
\begin{abstract}
  Translating characters instead of words or word-fragments has the
  potential to simplify the processing pipeline for neural machine
  translation (NMT), and improve results by eliminating
  hyper-parameters and manual feature engineering. However, it results
  in longer sequences in which each symbol contains less information,
  creating both modeling and computational challenges. In this paper,
  we show that the modeling problem can be solved by standard
  sequence-to-sequence architectures of sufficient depth, and that
  deep models operating at the character level outperform identical
  models operating over word fragments. This result implies that
  alternative architectures for handling character input are better
  viewed as methods for reducing computation time than as improved
  ways of modeling longer sequences. From this perspective, we
  evaluate 
  several techniques for character-level
  NMT, verify that they do not match the performance of our deep
  character baseline model, and evaluate the performance versus
  computation time tradeoffs they offer. Within this framework, we
  also perform the first evaluation for NMT of conditional computation
  over time, in which the model learns which timesteps can be skipped,
  rather than having them be dictated by a fixed schedule specified
  before training begins.
\end{abstract}

\section{Introduction}

Neural Machine Translation (NMT) has largely replaced the complex
pipeline of Phrase-Based MT with a single model that is trained
end-to-end. However, NMT systems still typically rely on pre- and
post-processing operations such as tokenization and word fragmentation
through byte-pair encoding~\citep[BPE; ][]{Sennrich2016}.
Although these are effective, they involve hyperparameters that should ideally be
tuned for each language pair and corpus, an expensive step that is
frequently omitted. Even when properly tuned, the
representation of the corpus generated by pipelined external processing
is likely to be sub-optimal. For instance, it is easy to find
examples of word fragmentations, such as {\em fling} $\rightarrow$ {\em fl} $+$
{\em ing}, that are linguistically implausible. NMT systems are generally
robust to such infelicities---and can be made more robust through subword regularization
\cite{Kudo2018}---but their effect on performance has not been
carefully studied. The problem of finding optimal segmentations becomes more
complex when an NMT system must handle multiple source and target languages, as
in multilingual translation or zero-shot approaches~\cite{Johnson2017}.

Translating characters instead of word fragments avoids these
problems, and gives the system access to all available information about source
and target sequences.
However, it presents significant modeling and
computational challenges. Longer sequences incur linear per-layer cost
and quadratic attention cost, and require information to be retained
over longer temporal spans. Finer temporal granularity also creates the potential
for attention jitter~\cite{Gulcehre2017}.
Perhaps most significantly, since the meaning of a word is not a
compositional function of its characters, the system must learn to
memorize many character sequences, a different task from the (mostly)
compositional operations it performs at higher levels of linguistic
abstraction.


In this paper, we show that a standard LSTM
sequence-to-sequence model works very well for characters, and given
sufficient depth, consistently outperforms identical models operating over word
fragments.
%
This result suggests that a productive line of research on character-level
models is to seek architectures that approximate standard sequence-to-sequence
models while being computationally cheaper. One approach to this problem is
{\em temporal compression}: reducing the number of state vectors required to
represent input or output sequences. We evaluate various approaches for
performing temporal compression, both according to a fixed schedule;
and, more ambitiously, learning compression decisions with a
Hierarchical Multiscale architecture \cite{Chung2017}. Following recent work by
\newcite{Lee2017}, we focus on compressing the encoder.

Our contributions are as follows:
\begin{itemize} \itemsep -0.1cm
\item The first large-scale empirical investigation of the translation
  quality of standard LSTM sequence-to-sequence architectures
  operating at the character level, demonstrating improvements in
  translation quality over word fragments,
  and quantifying the effect of corpus size and model capacity.

\item A comparison of techniques to compress character sequences,
  assessing their ability to trade translation quality for increased
  speed.
\item A first attempt to learn how to compress the source sequence
  during NMT training by using the Hierarchical Multiscale
  LSTM to dynamically shorten the source sequence
  as it passes through the encoder.
\end{itemize}

\section{Related Work}
Early work on modeling characters in NMT focused on solving
the out-of-vocabulary and softmax bottleneck problems associated with
word-level models~\cite{Ling2015, Costajussa2016, Luong2016}.
These took the form of word-boundary-aware hierarchical models, with
word-level models delegating to character-level models to generate
representations in the encoder and words in the decoder.
Our work will not assume fixed word boundaries are given in advance.

%
With the advent of word-fragment approaches, interest in
character-level processing fell off, but has recently been reignited with
the work of \newcite{Lee2017}.
They propose a specialized character-level encoder, connected
to an unmodified character-level RNN decoder.
They address the modeling and efficiency challenges of long character
sequences using a convolutional layer, max-pooling over time,
and highway layers.
%
%
We agree with their conclusion that character-level translation is
effective, but revisit the question of whether their specific encoder
produces a desirable speed-quality tradeoff in the context of a
much stronger baseline translation system.
We draw inspiration from their pooling solution for reducing sequence
length, along with similar ideas from the speech
community~\cite{Chan2016}, when devising fixed-schedule reduction
strategies in Section~\ref{sec:fixedStride}. 

One of our primary contributions is an extensive invesigation of the
efficacy of a typical LSTM-based NMT system when operating at the
character-level.
%
%
The vast majority of existing studies compare a specialized
character-level architecture to a distinct word-level one.
To the best of our knowledge, only a small number of papers have
explored running NMT unmodified on character sequences; these include:
\newcite{Luong2016} on WMT'15 English-Czech, \newcite{Wu2016} on
WMT'14 English-German, and \newcite{Bradbury2016} on IWSLT
German-English.
%
%
All report scores that either trail behind or reach parity with
word-level models.
Only \newcite{Wu2016} compare to word fragment models, which they
show to outperform characters by a sizeable margin.
%
We revisit the question of character- versus fragment-level NMT here,
and reach quite different conclusions.

\section{Methods}
\label{sec:methods}


\subsection{Baseline Sequence-to-Sequence Model}
\label{sec:baselineModel}

We adopt a simplified version of the LSTM architecture of
\newcite{Chen2018} that achieves state-of-the-art performance on the
competitive WMT14 English-French and English-German benchmarks. This
incorporates bidirectional LSTM (BiLSTM) layers in the encoder,
concatenating the output from forward and backward directions before
feeding the next layer.
Output from the top encoder layer is projected down to the decoder
dimension and used in an additive attention mechanism computed over the bottom
decoder layer.
The decoder consists of unidirectional layers, all of which use the
encoder context vectors computed from attention weights over the bottom layer.
For both encoder and decoder we use layer normalization~\cite{Ba2016}
and residual connections beginning at the third layer. We do not apply
a non-linearity to LSTM output. We regularize with dropout applied to
embeddings and to the output of each LSTM layer.

In the interests of simplicity and reproducibility, we depart 
from \newcite{Chen2018} in several ways:
we do not use multi-headed attention, 
feed encoder context vectors to the softmax,
regularize with label smoothing or weight decay, 
nor apply dropout to the attention mechanism.

Our baseline character models and BPE models both use this
architecture, differing only in whether the source and target languages are 
tokenized into sequences of characters or BPE word fragments. 
We describe BPE briefly below.


\subsection{Byte-Pair Encoding}
\label{sec:BPE}

Byte-Pair Encoding (BPE) offers a simple
interpolation between word- and character-level representations
\cite{Sennrich2016}.
%
It creates a vocabulary of frequent words and word fragments in an
iterative greedy merging process that begins with characters and ends
when a desired vocabulary size is reached.
The source and target language are typically processed together in
order to exploit lexical similarities.
Given a vocabulary, BPE re-tokenizes the corpus into word fragments in
a greedy left-to-right fashion, selecting the longest possible
vocabulary match, and backing off to characters when necessary.

%
Since each BPE token consists of one or more characters, BPE-tokenized
sequences will be shorter than character sequences.
Viewed as a mechanism to reduce sequence length,
BPE differs from the solutions we will discuss subsequently in that
it increases the vocabulary size, delegating the task of creating
representations for word fragments to the embedding table.
Also, despite being data-driven, its segmentation decisions are fixed
before NMT training begins.

\subsection{Fixed stride Temporal Pooling}
\label{sec:fixedStride}
%
%
We explore using fixed stride temporal pooling within the encoder
to compress the source character sequence.
These solutions are characterized by pooling the contents of two or
more contiguous timesteps to create a single vector that summarizes
them, and will replace them to shorten the sequence in the next layer.
These approaches can learn to interpret the raw character sequence
in service to their translation objective,
but any such interepretation
must fit into the pooling schedule that was specified during network
construction.
We evaluate two methods in this family: a re-implementation of
\newcite{Lee2017}, and a version of our baseline with
interspersed pooling layers.

%
%
As mentioned earlier, \newcite{Lee2017} propose a specialized
character encoder that combines convolutional layers to accumulate
local context, max-pooling layers to reduce sequence lengths, highway
layers to increase network capacity, followed by bidirectional GRU
layers to generate globally aware contextual source representations.
This strategy is particularly efficient because all reductions happen
before the first recurrent layer.
We re-implement their approach faithfully, with the exceptions of
using LSTMs in place of GRUs,\footnote{
Development experiments indicated that using LSTMs over GRUs resulted 
in a slight improvement.}
and modifying the batch sizes to accomodate our multi-GPU training scheme.

While pooling based approaches are typically employed in association with
convolutional layers, we can also intersperse pooling layers into our high
capacity baseline encoder.
This means that after each BiLSTM layer, we have the option to include
a fixed-stride pooling layer to compress the sequence before it is
processed by the next BiLSTM layer.
This is similar to the pyramidal LSTM encoders used for neural speech
recognition~\cite{Chan2016}.
This general strategy affords considerable flexibility to the network
designer, leaving the type of pooling (concatenation, max, mean),
and the strides with which to pool as design
decisions that can be tuned to fit the task.
%
%
%
%

\subsection{Learned Temporal Compression}
\label{sec:methodsLearned}
It is unsatisfying to compress a sequence on a fixed schedule;
after all, the characters in a sentence do not each carry an identical
amount of information.
%
%
The goal of this section is to explore data-driven reduction methods that are
optimized to the NMT system's objective, and which learn to compress
as a part of training.
%

Any strategy for performing temporal compression will necessarily make discrete
decisions, since sentence length is discrete. Examples of such strategies
include sparse attention~\cite{Raffel2017} and discrete
auto-encoders~\cite{Kaiser2018}. For our initial exploration, we chose the
hierarchical multiscale (HM) architecture of \newcite{Chung2017}, which we briefly
describe.


\subsubsection{Hierarchical Multiscale LSTM}
\label{sec:methodsHM}
The HM is a bottom-up temporal subsampling approach,
with each layer selecting the timesteps that will survive to the layer
above.
At a given timestep $t$ and layer $\ell$, the network makes
a binary decision, $z_t^\ell$, to determine whether or not it should
send its output up to layer $\ell+1$.
The preactivation for this decision, $\tilde{z}_t^\ell$, is a function
of the current node's inputs from below and from the previous hidden
state, similar to an LSTM gate.
However, $z_t^\ell$'s activation is a binary step function in the forward pass,
to enable discrete decisions, and a hard sigmoid in the backward pass, to allow
gradients to flow through the decision point.\footnote{This disconnect between
  forward and backward activations is known as a straight-through
  estimator~\cite{Bengio2013}.}
The $z_t^\ell$ decision affects both the layer above, and the next timestep of
the current layer:
\begin{itemize} \itemsep -0.05cm
\item $z_t^\ell=1$, \textit{flow up}: the node above ($t$, $\ell$+1)
  performs a normal LSTM update; the node to the right ($t$+1, $\ell$)
  performs a modified update called a \textit{flush}, which ignores 
  the LSTM internal cell at ($t$, $\ell$), and redirects the incoming
  LSTM hidden state from ($t$, $\ell$) to ($t$, $\ell+1$).
\item $z_t^\ell=0$, \textit{flow right}: the node above ($t$,
  $\ell$+1) simply copies the cell and hidden state values from
  ($t-1$, $\ell$+1); the node to the right \mbox{($t$+1, $\ell$)}
  performs a normal LSTM update.
\end{itemize}
Conceptually, when $z_t^\ell=0$, the node above it becomes a
placeholder and is effectively removed
from the sequence for that layer.
Shorter upper layers save computation and facilitate the
left-to-right flow of information for the surviving nodes.

%
Typically, one uses the top hidden state $h_t^L$ from a stack of $L$ RNNs
to provide the representation for a timestep $t$.
But for the HM, the top layer may be updated much less frequently
than the layers below it.
To enable tasks that need a distinct representation for each timestep,
such as language modeling,
the HM employs a \textit{gated output module} to mix hidden states across 
layers.
This learned module combines the states \mbox{$h_t^1$, $h_t^2$, $\ldots$, $h_t^L$}
using scaling and projection operators to produce a single output $h_t$.

\subsubsection{Modifying the HM for NMT}
%

We would like sequences to become progressively shorter
as we move upward through the layers.
%
As originally specified, the HM calculates $z_t^\ell$
independently for every $t$ and $\ell$, including copied nodes,
meaning that a ``removed'' timestep could reappear in a higher
layer when a copied node ($t$, $\ell$) sets $z_t^\ell=1$.
This is easily addressed by locking $z_t^\ell=0$ for copied nodes,
creating a hierarchical structure in which upper layers never
increase the amount of computation.
%

We also found that the flush component of the original architecture,
which modifies the LSTM update at ($t$+1, $\ell$) to discard the LSTM's
internal cell, provided too much incentive to leave $z_t^\ell$ at $0$,
resulting in degenerate configurations which collapsed to having very few
tokens in their upper layers.
%
We addressed this by removing the notion of a flush from our
architecture. The node to the right ($t$+1, $\ell$) always performs a
normal LSTM update, regardless of $z_t^\ell$.
This modification is similar to one proposed independently by 
\newcite{Kadar2018}, who simplified the flush operation
by removing the connection to ($t$, $\ell+1$).

We found it useful to change the initial value of the bias
term used in the calculation of $\tilde{z}_t^\ell$, which we refer to
as the $z$-bias. Setting $z$-bias to 1, which is the saturation point
for the hard sigmoid with slope 1, improves training stability by encouraging
the encoder to explore configurations where most timesteps survive through all layers,
before starting to discard them.
%
%

Even with these modifications, we observed degenerate behavior in some
settings. To discourage this, we added a compression loss component
similar to that of \newcite{Ke2018} to penalize $z$ activation rates
outside a specified range $\alpha_1,\alpha_2$: $\mathcal{L}_c = \sum_l
\max(0,\; Z^l-\alpha_1 T,\; \alpha_2 T-Z^l)$, where 
$T$ is source sequence length and
$Z^l = \sum_{t=1}^T z_t^l$.

To incorporate the HM into our NMT encoder, we replace the lowest
BiLSTM layer with unidirectional HM layers.\footnote{The flush operation 
  makes the original HM
  inherently left-to-right. Since we have dropped flushes from our
  current version, it should be straightforward to devise a
  bidirectional variant, which we leave to future work.}  
We adapt any remaining BiLSTM layers to copy or
update according to the z-values calculated by the top HM layer.

\section{Experimental Design}

\subsection{Corpora}

We adopt the corpora used by Lee et al \shortcite{Lee2017}, with the
exception of WMT15 Russian-English.\footnote{Due to licence restrictions.}
To measure performance on an ``easy'' language pair, and
to calibrate
our results against recent benchmarks, we also included WMT14
English-French. Table~\ref{tab:corpora} gives details of the corpora used.
All corpora are preprocessed using Moses tools.\footnote{
  Scripts and arguments:\\
  \tt
  remove-non-printing-char.perl \\
  tokenize.perl\\
  clean-corpus-n.perl -ratio 9 1 100
}
Dev and test corpora
are tokenized, but not filtered or cleaned.
Our character models use only the most frequent 496 characters across both source and target languages;
similarly, BPE is run across both languages, with a vocabulary size of 32k.

\begin{table}[t]
\begin{tabular}{l|rll}
  corpus & train & dev & test \\
  \hline
  WMT15 Finnish-En & 2.1M & 1500 & 1370 \\
  WMT15 German-En & 4.5M & 3003 & 2169 \\
  WMT15 Czech-En & 14.8M & 3003 & 2056 \\
  WMT14 En-French & 39.9M & 3000 & 3003 \\
\end{tabular}
\caption{Corpora, with
linecounts. Test sets are WMT14-15
newstest.  Dev sets are newsdev 2015 (Fi) and newstest 2013
  (De, Fr), and 2014 (Cs).}
\label{tab:corpora}
\end{table}

\subsection{Model sizes, training, and inference}

Except where noted below, we used 6 bidirectional layers in the
encoder, and 8 unidirectional layers in the decoder. All
vector dimensions were 512.

Models were trained using sentence-level cross-entropy loss. Batch sizes are
capped at 16,384 tokens, and each batch is divided among 16 NVIDIA P100s
running synchronously.

Parameters were initialized with a uniform (0.04) distribution.
We use the Adam optimizer, with $\beta_1$ = $0.9$, $\beta_2$ = $0.999$, and
$\epsilon$ = $10^{-6}$ \cite{Kingma2014}. Gradient norm is clipped to
5.0. The initial learning rate is 0.0004, and we halve it
whenever dev set perplexity has not decreased for 2k batches, with at
least 2k batches between successive halvings. Training stops when
dev set perplexity has not decreased for 8k batches.

Inference uses beam search with 8 hypotheses,
coverage penalty of 0.2 \cite{tu2016modeling}, and length normalization of
0.2 \cite{Wu2016}.

\subsection{Tuning and Evalution}

When comparing character-level and BPE models, we tuned dropout
independently for each setting, greedily exploring increments of 0.1
in the range 0.1--0.5, and selecting based on dev-set BLEU.
This expensive strategy is crucial to obtaining
valid conclusions, since optimal dropout values tend to be lower for
character models.

%


Our main evaluation metric is Moses-tokenized case-sensitive BLEU
score. We report test-set scores on the checkpoints having highest dev-set BLEU.
To facilitate comparison with future work we also report SacreBLEU scores
\cite{Post18} for key results, using the Moses detokenizer.



\section{Results}

\subsection{Character-level translation}
\label{sec:ExpBPEvsChar}

\begin{table}[t]
\centering
\begin{tabular}{l|ccc|c}
& \multicolumn{3}{|c|}{Tokenized BLEU} & SacreBLEU \\
Language & BPE & Char & Delta & Char\\
\hline
EnFr & 38.8 & 39.2 & 0.4 & 38.1\\
CsEn & 24.8 & 25.9 & 1.1 & 25.6\\
DeEn & 29.7 & 31.6 & 1.9 & 31.6\\
FiEn & 17.5 & 19.3 & 1.8 & 19.5\\
\end{tabular}
\caption{Character versus BPE translation.}
\label{tab:CharVsWpm}
\end{table}
\begin{table}[t]
\centering
\begin{tabular}{llcc|c}
\multicolumn{3}{l}{Comparison Point} & Ref & Ours \\
\hline
\newcite{Chen2018} & BPE & EnFr  & 41.0 & \multirow{ 2}{*}{38.8} \\
\newcite{Wu2016}   & BPE & EnFr  & 39.0 & \\
\newcite{Lee2017} & Char & CsEn  & 22.5 & 25.9 \\
                  &      & DeEn  & 25.8 & 31.6 \\
                  &      & FiEn  & 13.1 & 19.3 \\
\end{tabular}
\caption{Comparisons with some recent points in
  the literature. Scores are tokenized BLEU.}
\label{tab:otherPapers}
\end{table}

We begin with experiments to compare the standard RNN architecture
from Section~\ref{sec:baselineModel} at the character and BPE levels,
using our full-scale model with 6 bidirectional encoder layers and 8
decoder layers.
The primary results of our experiments are presented in
Table~\ref{tab:CharVsWpm}, while Table~\ref{tab:otherPapers} positions
the same results with respect to recent points from the literature.

There are a number of observations we can draw from this data.
First, from the EnFr results in Table~\ref{tab:otherPapers},
we are in line with GNMT~\cite{Wu2016}, and within 2 BLEU points of
the RNN and Transformer models investigated by \newcite{Chen2018}.
So, while we are not working at the exact state-of-the-art, we
are definitely in a range that should be relevant to most
practitioners.

Also from Table~\ref{tab:otherPapers},
we compare quite favorably with \newcite{Lee2017}, exceeding their reported scores by
3-6 points, which we attribute to having employed much higher model
capacity, as they use a single bidirectional layer in the encoder
and a two-layer decoder.
We investigate the impact of model capacity in
Section~\ref{sec:ExpCapacity}.

Finally, Table~\ref{tab:CharVsWpm} clearly shows the character-level
systems outperforming BPE for all language pairs.
The dominance of character-level methods in Table~\ref{tab:CharVsWpm}
indicates that RNN-based NMT architectures are not only
capable of translating character sequences, but actually benefit from
them.
This is in direct contradiction to the few previously reported results
on this matter, which can in most cases be explained by our increased
model capacity.
The exception is GNMT~\cite{Wu2016}, which had similar depth.
In this case, possible explanations for the discrepancy include our use
of a fully bidirectional encoder, our translating into English instead of
German, and our model-specific tuning of dropout.

\subsubsection{Effect of model capacity}
\label{sec:ExpCapacity}
\begin{figure*}[t]
  \centering
    \includegraphics[width=0.4\textwidth]{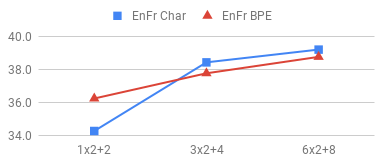}
    \includegraphics[width=0.4\textwidth]{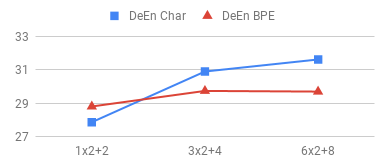}
    \includegraphics[trim=0 0.6cm 0 0, width=0.4\textwidth]{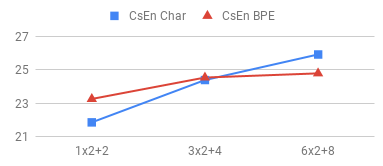}
    \includegraphics[trim=0 0.6cm 0 0, width=0.4\textwidth]{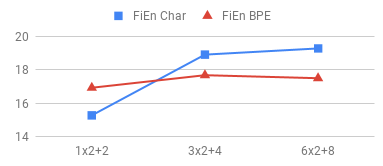}
  \caption{Test BLEU for character and BPE translation as
    architectures scale from 1 BiLSTM encoder layer and 2 LSTM decoder
    layers (1$\times$2$+$2) to our standard 6$\times$2$+$8.  The y-axis
    spans 6 BLEU points for each language pair.}
  \label{fig:BleuVsArch}
\end{figure*}

Character-level NMT systems have a more difficult sequence-modeling task,
as they need to infer the meaning of words from their constituent
characters, where models with larger tokens instead delegate this
task to the embedding table.
Therefore, we hypothesize that increasing the model's capacity by
adding layers will have a greater impact on character-level models.
Figure~\ref{fig:BleuVsArch} tests this hypothesis by measuring the
impact of three model sizes on test BLEU score.
%
For each of our four language pairs, the word-fragment model starts out
ahead, and quickly loses ground as architecture size increases.
For the languages with greater morphological complexity---German, Czech
and Finnish---the slope of the character model's curve is notably
steeper than that of the BPE system, indicating that these
systems could benefit from yet more modeling capacity.

\subsubsection{Effect of corpus size}
\label{sec:ExpCorpusSize}
\begin{figure}[t]
  \centering
  \includegraphics[trim=0 0.6cm 0 0, width=0.45\textwidth]{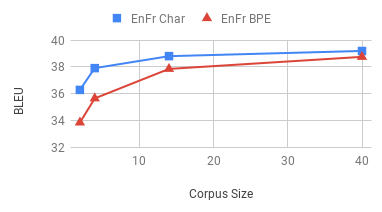}
  \caption{BLEU versus training corpus size in millions of sentence
    pairs, for the EnFr language-pair.}
  \label{fig:BleuVsCorpus}
\end{figure}

One of the most compelling arguments for working with characters (and
to a lesser extent, word-fragments) is improved generalization.
Through morphological generalizations, the system can better handle
low-frequency and previously unseen words.
It stands to reason that as the training corpus increases in size, the
importance of these generalization capabilities will decrease.
We test this hypothesis by holding the language pair constant, and
varying the training corpus size by downsampling the full training corpus.
We choose EnFr because it has by far the most available
data.
We compare four sizes:
2M, 4M, 14M and 40M.

The results are shown in Figure~\ref{fig:BleuVsCorpus}.
As expected, the gap between character and word-fragment modeling
decreases as corpus size increases.
From the slopes of the curves, we can infer that the advantage of
character-level modeling will disappear completely as we reach 60-70M
sentence pairs.
However, there is reason to expect this break-even point
to be much higher for more morphologically complex languages.
It is also important to recall that relatively few language-pairs can assemble
parallel corpora of this size.

\subsubsection{Speed}
\label{sec:Speed}

\begin{figure}[t]
  \centering
  \includegraphics[trim=0 0.6cm 0 0, width=0.45\textwidth]{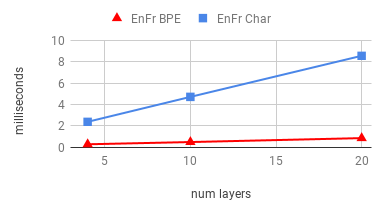}
  \caption{Training time per sentence versus total number of layers (encoder plus decoder) in
    the model.} 
  \label{fig:timing}
\end{figure}

The performance advantage of working with characters comes at a significant
computational cost.
With our full-sized
architecture, character models trained roughly 8x more slowly than BPE models.%
\footnote{Recall that we use batches
containing 16,384 tokens---corresponding to a fixed memory budget---for both
character and BPE models. Thus character models are slowed not only by having
longer sentences, but also by parallelizing across fewer sentences in each batch.}
Figure~\ref{fig:timing} shows that training time grows
linearly with number of layers in the model, and that character models
have a much higher per-layer cost: roughly 0.38 msec/sentence versus 0.04 for
BPE.
We did not directly measure the difference in attention cost, but it
cannot be greater than the difference in total cost for the smallest
number of layers.
Therefore, we can infer from Figure~\ref{fig:timing}
that processing 5 layers in a character model incurs roughly the same
time cost as attention. 
This is surprising given the quadratic cost of attention, and
indicates that efforts to speed up character models cannot focus
exclusively on attention.



\subsubsection{Qualitative comparison}

To make a qualitative comparison between word fragments (BPE) and 
characters for NMT, we examined 100 randomly selected sentence pairs from 
the DeEn test set.
One author examined the sentences, using a display that showed the 
source\footnote{The annotating author does not speak German.}
and the reference, along with the output of BPE and character models.
%
Any differences between the two outputs were highlighted.
They then assigned tags to both system outputs indicating broad error 
categories, such as lexical choice, word order and German compound 
handling.%
\footnote{Our annotator also looked specifically for  
agreement and negation errors, as studied by \newcite{Sennrich2017}
for English-to-German character-level NMT.
However, neither system exhibited these error types with sufficient 
frequency to draw meaningful conclusions.
}
Tags were restricted to cases where one system made a mistake that the 
other did not.

Of the 100 sentences, 47 were annotated as being identical or of roughly 
the same quality.
The remaining 53 exhibited a large variety of differences.
Table~\ref{tab:errorAnalysis} summarizes the errors that were most easily 
characterized.
BPE and character systems differ most in the number of lexical choice 
errors, and in the extent to which they drop content.
The latter is surprising, and appears to be a side-effect of 
a general tendency of the character models to be more faithful to the 
source, verging on being overly literal.
An example of dropped content is shown in Table~\ref{tab:errorExamples} 
(top).

Regarding lexical choice, the two systems differ not only in the number 
of errors, but in the nature of those errors.
In particular, the BPE model had more trouble handling German compound 
nouns.
Table~\ref{tab:errorExamples} (bottom) shows an example which exhibits 
two compound errors, including
one where the character system is a strict improvement, translating 
\textit{Bunsenbrenner} into \textit{bunsen burner} instead of 
\textit{bullets}.
The second error follows another common pattern, where 
both systems mishandle the German compound (\textit{Chemiestunden / 
chemistry lessons}), but the character system fails in a more useful way.

We also found that both systems occasionally mistranslate proper names.
Both fail by attempting to translate when they should copy over, 
but the BPE system's errors are harder to understand as they involve 
semantic translation, rendering \textit{Britta Hermann} as 
\textit{Sir Leon}, and \textit{Esme Nussbaum} as \textit{smiling 
walnut}.\footnote{
The BPE segmentations for these names were: 
\texttt{\detokenize{_Britt a _Herr mann}} and  
\texttt{\detokenize{_Es me _N uss baum}}}
The character system's one observed error in this category was phonetic 
rather than semantic, rendering \textit{Schotten} as 
\textit{Scottland}.

Interestingly, we also observed several instances where the model
correctly translates the German 24-hour clock into the English 12-hour 
clock; for example, \textit{19.30} becomes \textit{7:30 p.m.}.
This deterministic transformation is potentially in reach for both 
models, but we observed it only for the character system in this sample.

\begin{table}[t]
    \centering
    \begin{tabular}{lrr}
        Error Type                  & BPE & Char \\ 
        \hline
        Lexical Choice              &  19 &    8 \\
        \hspace*{1em}Compounds      &  13 &    1 \\
        \hspace*{1em}Proper Names   &   2 &    1 \\
        \hspace*{1em}Morphological  &   2 &    2 \\
        \hspace*{1em}Other lexical  &   2 &    4 \\
        Dropped Content             &   7 &    0 \\
    \end{tabular}
    \caption{Error counts out of 100 randomly sampled examples from the 
             DeEn test set.}
    \label{tab:errorAnalysis}
\end{table}
\begin{table*}[t]
    \centering
    \begin{tabular}{l p{14.2cm}}
        \hline
        Src  & Für diejenigen, die in ländlichen und abgelegenen Regionen \textbf{des Staates} lebten, $\ldots$ \\
        Ref  & Those living in regional and remote areas \textbf{of the state} $\ldots$ \\
        BPE  & For those who lived in rural and remote regions, $\ldots$ \\
        Char & For those who lived in rural and remote regions \textbf{of the state}, $\ldots$ \\
        \hline
        Src & Überall im Land, in Tausenden von \textbf{Chemiestunden}, 
              haben Schüler ihre \textbf{Bunsenbrenner} auf 
              Asbestmatten abgestellt.\\
        Ref & Up and down the country, in myriad \textbf{chemistry 
              lessons}, pupils have perched their \textbf{Bunsen burners} 
              on asbestos mats. \\
        BPE & Across the country, thousands of \textbf{chemists} have 
              turned their \textbf{bullets} on asbestos mats. \\
        Char & Everywhere in the country, in thousands of \textbf{chemical
              hours, students} have parked their \textbf{bunsen 
              burners} on asbestos mats. \\
        \hline
    \end{tabular}
    \caption{Examples of BPE and character outputs for two sentences from the DeEn test set, 
    demonstrating dropped content (top) and errors with German compounds (bottom).}
    \label{tab:errorExamples}
\end{table*}

\subsection{Compressing the Source Sequence}

At this point we have established that character-level NMT benefits translation
quality, but incurs a large computational cost. In this section, we evaluate the
speed-quality tradeoffs of various techniques 
for reducing the number of state vectors required to represent the source sentence.
All experiments are conducted on our DeEn language pair, 
chosen for having a good balance of morphological complexity
and training corpus size.

\subsubsection{Optimizing the BPE vocabulary}
\label{sec:ExpBPECompression}

Recall that BPE interpolates between word- and character-level processing
by tokenizing consecutive characters into word fragments;
larger BPE vocabulary sizes result in larger fragments and shorter sequences. 
If character-level models outperform BPE
with a vocabulary size of 32k, then is there a smaller BPE vocabulary size that
reaps the benefits of character-level processing, while still substantially
reducing the sequence length?

To answer this question, we test a number of BPE vocabularies, 
as shown in Table~\ref{tab:fixedcomp}. For each vocabulary, we
measure BLEU and sequence compression rate, defined as the average size of the
source sequence in characters divided by its size in word fragments (the ratio
for the target sequence was similar). 
%
%
Unfortunately, even at just 1k vocabulary items,
BPE has already lost a BLEU point with respect to the character model.
%
When comparing these results to the
other methods in this section, it is important to recall that BPE is
compressing both the source and target sequence (by approximately the same
amount), doubling its effective compression rate.

\subsubsection{Fixed Stride Compression}
\label{sec:ExpFixedStride}
\begin{table}[t]
\centering
\begin{tabular}{l|lll}
Encoder            & BPE Size & BLEU  & Comp.           \\ \hline
BiLSTM             & Char     & 31.6   & 1.00            \\
BiLSTM             & 1k       & 30.5   & 0.44            \\
BiLSTM             & 2k       & 30.4   & 0.35            \\
BiLSTM             & 4k       & 30.0   & 0.29            \\
BiLSTM             & 8k       & 29.6   & 0.25            \\
BiLSTM             & 16k      & 30.0   & 0.22            \\
BiLSTM             & 32k      & 29.7   & 0.20            \\\hline
Lee et. al. reimpl & Char     & 28.0   & 0.20            \\
BiLSTM + pooling   & Char     & 30.0   & 0.47             \\
\hline
HM, 3-layer        & Char     & 31.2   & 0.77 \\ 
HM, 2-layer        & Char     & 30.9   & 0.89    
\end{tabular}
\caption{Compression results on WMT15 DeEn. The
  \textit{Comp.} column shows the ratio of total computations carried out in the
  encoder.}
\label{tab:fixedcomp}
\end{table}
%
The goal of these experiments is to determine whether using fixed schedule
compression is a feasible alternative to BPE.
%
We evaluate our re-implementation of the pooling model of
\newcite{Lee2017} and our pooled BiLSTM encoder,
both described in Section~\ref{sec:fixedStride}.
%
%
For the pooled BiLSTM encoder, 
development experiments led us to introduce two mean-pooling layers, 
a stride 3 layer after the second BiLSTM, 
and a stride 2 layer after the third.
Therefore, the final output of the encoder is compressed by a factor
of 6. 
%
%
%

The results are also shown in Table~\ref{tab:fixedcomp}.
Note that for the pooled BiLSTM, different encoder layers have different
lengths: 2 full length layers, followed by 1 at $\frac{1}{3}$ length 
and 3 at $\frac{1}{6}$ length.
Therefore, we report the average compression across layers here
and for the HM in Section~\ref{sec:ExpHM}.

Our implementation of \newcite{Lee2017} outperforms the original results by
more than 2 BLEU points. We suspect most
of these gains result from better optimization of the model with large batch
training. However, our attempts to scale this encoder to larger depths,
and therfore to the level of performance exhibited by our other systems,
did not result in any significant improvements.
This is possibly due to difficulties with optimizing a deeper stack of diverse layers.

Comparing the performance of our Pooled BiLSTM model against BPE,
we notice that for a comparable level of compression (BPE size of 1k), BPE
out-performs the pooled model by around 0.5 BLEU points. 
At a similar level of performance (BPE size of 4k), BPE has significantly shorter
sequences.
%
Although fixed-stride pooling does not yet match the performance of BPE,
we remain optimistic about its potential.
The appeal of these models derives from their simplicity; they
are easy to optimize, perform reasonably well, and remove the
complication of BPE preprocessing.

\subsubsection{Hierarchical Multiscale Compression}
\label{sec:ExpHM}
\begin{table}[t]
\centering
\begin{tabular}{l|rr}
Model & BLEU & Comp. \\
\hline
LSTM                         & 28.9 & 1.00  \\ 
HM, no-fl                    & 27.3 & 0.63  \\ 
HM, no-fl, hier              & 28.5 & 0.65  \\ 
HM, no-fl, hier, zb1, anneal & 28.8 & 0.65  \\ 
\hline
\end{tabular}
\caption{HM small-scale results on WMT15 DeEn. The {\em Comp.} column
  is the proportion of layer-wise computation relative to the full
  LSTM.}
\label{tab:HMResults}
\end{table}

We experimented with using the Hierarchical Multiscale (HM; Section~\ref{sec:methodsHM})
architecture to learn compression decisions for the encoder.

For initial exploration, we used a scaled-down architecture consisting
of 3 unidirectional HM encoder layers and 2 LSTM decoder layers, 
attending over the HM's gated output module.
%
Comparisons to an equivalent LSTM are shown in
table~\ref{tab:HMResults}.
The first two HM lines justify the no-flush and hierarchical
modifications described in Section~\ref{sec:methodsHM}, yielding
incremental gains of 27.3 (the flush variant failed to converge), and 1.2
respectively. Initializing z-bias to 1 and annealing the slope of the
hard binarizer from 1.0 to 5.0 over 80k minibatches gave further small
gains, bringing the HM to parity with the LSTM while saving
approximately 35\% of layer-wise computations. Interestingly, we found
that, over a wide range of training conditions, each layer tended to
reduce computation by roughly 60\% relative to the layer
below.\footnote{For instance, the 2nd and 3rd layer of the best
  configuration shown had on average 60\% and 36\% of $z$ gates open,
  yielding the computation ratio of $(1 + 0.6 + 0.36) / 3 = 0.65$.}
  
For full-scale experiments, we stacked 5 BiLSTM layers on top of 2 or 3 HM
layers, as described in section~\ref{sec:methodsHM}, using only the top HM layer
(rather than the gated output module) as input to the lowest BiLSTM layer.
To stabilize the 3-HM configuration we used a compression penalty with a
weight of 2, and $\alpha_1$ and $\alpha_2$ of 0.1 and 0.9.
Given the tendency of HM
layers to reduce computation by a roughly constant proportion, we
expect fewer z-gates to be open in the 3-HM configuration, but this is
achieved at the cost of one extra layer relative to our standard
12-layer encoder. 
%
%
As shown in table~\ref{tab:fixedcomp}, the 3-HM configuration achieves much better
compression even when this is accounted for, and also gives slightly better
performance than 2-HM.
In general, HM gating results in less
compression but better performance than the
fixed-stride techniques.

Although these preliminary results are promising, it should be emphasized that the speed
gains they demonstrate are conceptual, and that realizing them in practice
comes with significant engineering challenges.

\section{Conclusion}
We have demonstrated the translation quality of standard NMT
architectures operating at the character-level.
%
%
Our experiments show the surprising result that character NMT can
substantially out-perform BPE tokenization for all but the largest
training corpora sizes, and the less surprising result that doing so
incurs a large computational cost.
To address this cost, we have explored a number of methods for source-sequence
compression, including the first application of the Hierarchical Multiscale
LSTM to NMT, which allows us to learn to dynamically compress the source
sequence.

We intend this paper as a call to action.
Character-level translation is well worth doing, but we do not yet
have the necessary techniques to benefit from this quality boost
without suffering a disproportionate reduction in speed.
We hope that these results will spur others to revisit the question of
character-level translation as an interesting test-bed for methods
that can learn to process, summarize or compress long sequences.

\bibliography{emnlp2018} \bibliographystyle{acl_natbib_nourl.bst}

\end{document}